\documentclass[numbered]{trbunofficial}
\usepackage{lineno}
\usepackage[hidelinks]{hyperref}
\usepackage{graphicx}
\usepackage{algorithm} 
\usepackage{algpseudocode}
\usepackage{caption}
\usepackage{subcaption}

\AuthorHeaders{Li, Fischer, Kontar, and Ahn}
\title{Predicting Vehicles' Longitudinal Trajectories and Lane Changes on Highway On-Ramps}

\author{
  \textbf{Nachuan Li$^{*}$}\\
  University of Wisconsin-Madison Department of Civil and Environmental Engineering\\
  \hfill\break
   \textbf{Riley Fischer}\\
  University of Wisconsin-Madison Department of Mathematics\\
  \hfill\break
  \textbf{Wissam Kontar}\\
  University of Wisconsin-Madison Department of Civil and Environmental Engineering\\
  \hfill\break
  \textbf{Soyoung Ahn, PhD}\\
  University of Wisconsin-Madison Department of Civil and Environmental Engineering
 }

\begin{document}
\maketitle

\section{Abstract}
Vehicles on highway on-ramps are one of the leading contributors to congestion. In this paper, we propose a prediction framework that predicts the longitudinal trajectories and lane changes (LCs) of vehicles on highway on-ramps and tapers. Specifically, our framework adopts a combination of prediction models that inputs a 4 seconds duration of a trajectory to output a forecast of the longitudinal trajectories and LCs up to 15 seconds ahead. Training and Validation based on next generation simulation (NGSIM) data show that the prediction power of the developed model and its accuracy outperforms a traditional long-short term memory (LSTM) model. Ultimately, the work presented here can alleviate the congestion experienced on on-ramps, improve safety, and guide effective traffic control strategies.

\hfill\break
\noindent\textit{Keywords}: On-ramp, Car-following Model, Mandatory Lane Change, Trajectory Prediction, LSTM, Random Forest

\newpage

\section{Introduction}

\subsection{Overview}

With an increased number of vehicles in the recent years, congestion on highways has become a serious problem. One of the main location of highway congestion and breakdown is the merging sections, including the on-ramps and the tapers \cite{cassidy1999some,yang1996microscopic, ahmed1999modeling}. Also, the motions of vehicles on highway on-ramps are complicated, because they entail both car-following (CF) and LC behaviors. In particular, the type of LC behavior from an on-ramp to the main stream of a highway is called mandatory lane change (MLC), which is a kind of LC guided by a vehicle's macroscopic routing decision and has generally a lower gap acceptance due to urgency \cite{yang1996microscopic, cassidy1999some,ahmed1999modeling}. 

Another challenge of analysing vehicle trajectories at highway on-ramps is the lack of data. While a lot of studies have used vehicle trajectory data sets for vehicle trajectory analysis, most of them focus on a basic freeway segment, which has more trajectories available and does not involve MLCs \cite{wu2017multi,suh2018stochastic, dou2019gated, altche2017lstm, 6648406}. For example, in NGSIM data set, there are at least 6101 trajectories collected for US Highway 101 \cite{altche2017lstm}, yet only 397 are of them are those of on-ramp vehicles. The lack of trajectories make training and testing particularly difficult. Therefore, the trajectories of on-ramp vehicles is research-worthy.

In this paper, we propose a prediction framework to predict the longitudinal positions and LCs of vehicles on highway on-ramps. To specify how our model works, we partition the model into the following steps:\\
\begin{enumerate}
    \item LSTM networks to pre-train the longitudinal positions of the neighboring vehicles
    \item A CF model to fit a central vehicle's accelerations of the initial 4 seconds of its trajectory
    \item Random Forest (RF) classifiers to predict the LCs in the next 15 seconds given the last time step of the initial data.\\
\end{enumerate}  
Given 4 seconds of initial trajectories, we predict the longitudinal trajectories and LCs in the next 15 seconds. Then, we will use this Model to train and validate the on-ramp vehicles' trajectories with Interstate-80 Freeway (I80) data set from NGSIM. The reason why we use a very short initial trajectory is because the on-ramp is relatively short, and it takes on average less than 25 seconds for vehicles to travel through. We ensure there is a long enough trajectory to perform forecasts and validations.

\subsection{Related Works}

An overview of literature suggests that CF behaviors of vehicles and prediction of their longitudinal trajectories is a widely studied topic. A number of studies have analyzed and calibrated the existing CF models for an effective simulation or motion estimation \cite{olstam2004comparison, treiber2000congested, rakha2003comparison, aycin1999comparison, chen2020modeling}. For example, Olstam and Tapani \cite{olstam2004comparison} have compared the CF models in AIMSUN, MITSIM, VISSIM, and Fritzsche micro-simulation packages on reaction magnitudes and reaction times of central vehicles. Chen, Wang, and Dong \cite{chen2020modeling} have extracted trajectory data from a highway on-ramp in Nanjing, China, and have applied an Intelligent Driver Model (IDM) to fit the vehicles' accelerations. Other studies have applied recurrent neural networks to forecast the trajectory of vehicles \cite{altche2017lstm, kim2017probabilistic, ye2016vehicle}. For example, Altche and de La Fortelle \cite{altche2017lstm} have applied a LSTM neural network to forecast the trajectories of vehicles in NGSIM US Highway 101 data set. They treat the first 10 seconds of trajectory of each central vehicle and its neighbors as inputs and forecast the trajectories of each central vehicles for the next 10 seconds, and have got promising accuracy for both longitudinal and lateral position forecasts \cite{altche2017lstm}.

MLCs of vehicles are also widely studied. In general, a gap acceptance model is widely used for modeling MLCs \cite{doi:10.3141/2188-12,ali2018connectivity, hao2020research}. According to this model, a lower speed, a higher density on the target lane, and a closer position to the end of the current lane, all lead to lower gap acceptance \cite{doi:10.3141/2188-12,ali2018connectivity, hao2020research}. According to Hao et al. \cite{hao2020research}, the minimum merging headway is proportional to the remaining distance during which a MLC must be completed. In addition, Machine learning classifiers is also widely applied for predicting MLCs\cite{dou2019gated, 6648406, liu2019deep}. For example, Dou et al. \cite{dou2019gated} have applied a gated branch neural network to predict the LCs of vehicles at an on-ramp. Through validation on NGSIM data set, they got an accuracy of over 96\%, which is higher than any other existing model. While not specifically used for only MLCs, stochastic models are also widely applied to model LCs \cite{wang2019state, zhang2018addressing, suh2018stochastic}. For example, Wang, Zhang, and Jiao \cite{wang2019state} have used a hidden Markov model to characterize MLCs of vehicles and have got a high accuracy when testing with NGSIM data set.

To sum up, a large number of existing studies have analyzed CF and MLC behaviors, or predicted the trajectories of vehicles on the highway. In addition, NGSIM data set is widely used for validation. However, most of the CF models fail to address the contributions of the neighbors on the adjacent lane to a vehicle's acceleration. Also, based on our knowledge, there is no research so far dedicated to predicting the trajectories at a highway on-ramp, most likely due to the lack of trajectory data sets. Therefore, the trajectories of vehicles at an on-ramp is relatively under-explored.

In the rest of this paper, we will introduce the definitions and notations used, our proposed prediction framework, the training and testing data set, and the accuracy results we get.\\

\section{Definitions and Notations}
Before discussing our prediction framework in details, it is necessary to introduce some definitions and notations used in the rest of the paper. To avoid confusion, we use the word "pre-trained" for results derived from a LSTM model, "forecasted" for results derived from a CF model, and "classified" for results derived from a RF classifier.

\begin{figure}
    \centering
    \includegraphics[width=0.8\linewidth]{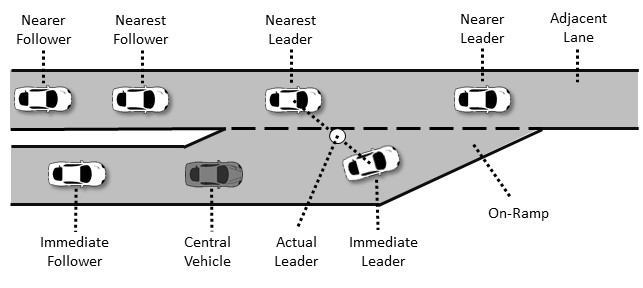}
    \caption{Geometry of a Central Vehicle and its Neighbors}
    \label{vm}
\end{figure}

Fig.\ref{vm} shows the relative locations of the neighboring vehicles with respect to an on-ramp vehicle. In the rest of the paper, a central vehicle refers to an on-ramp vehicle whose trajectory and LC is going to be forecasted. Then, we denote the closest leader and the closest follower on the same lane as the immediate leader and the immediate follower respectively. We denote the lane the central vehicle is going to merge into as the adjacent lane. There are four important vehicles on the adjacent lane: the nearest leader, the nearest follower, the nearer leader, and the nearer follower. Finally, we use the term actual leader to refer to the virtual vehicle that is in the geometric center between the immediate leader and the nearest leader if both of the two vehicles exist. In Fig.\ref{vm}, the actual leader is indicated by the white circle. However, in situations where the immediate leader is missing because the central vehicle is too close to the end of the taper, we set the actual leader as the nearest leader.

Then we introduce some common notations used in the paper. First, it is important to introduce the meaning of the subscripts of parameters. Eq.\ref{n_val} shows the letter notation taken for a subscript $n$ for different vehicles.\\

\begin{equation}\label{n_val}
\centering
n :=
    \begin{cases}
   i, &\text{the central vehicle}\\ 
    l, &\text{the immediate leader} \\
    f, &\text{the immediate follower} \\
    l1, &\text{the nearest leader}\\
    l2, &\text{the nearer leader}\\
    f1, &\text{the nearest follower}\\
    f2, &\text{the nearer follower}\\
    m, &\text{the actual leader}
    \end{cases}
\end{equation}\\

Table \ref{table:lc_4veh} shows the description of each notation for an arbitrary subscript $n$. Also, we denote $x_\text{end}$ as the longitudinal position of the end of the on-ramp, $y_\text{cur}$ and $y_\text{tar}$ as the lateral positions of the lane center of the current lane and the target lane respectively, and $\Omega_{n}$ as the set of information of a vehicle $n$, which includes its position, speed and acceleration, as defined by Eq.\ref{veh}. \\

\begin{equation}\label{veh}
    \Omega_n:=\{x_n, y_n, v_n, u_n, a_n, e_n\}
\end{equation}\\

Finally, we introduce the usage of accents. For an arbitrary parameter $X$, $\tilde{X}$ denotes the pre-trained value with an LSTM model, and $\hat{X}$ denotes the forecasted value with a CF model. For example, $\tilde{\Omega}_l$ is the set of pre-trained information of the leader, and $\hat{v}_i$ is the forecasted velocity of the central vehicle, 

\begin{center}\label{lc_4veh}
\begin{table}[]
\centering
\begin{tabular}{p{2.5cm}  p{12cm} }
\hline\hline
\textbf{Notation} & \textbf{Description} \\
\hline\hline
$x_n$ & The longitudinal position of a vehicle $n$ in m \\
\hline
$y_n$ & The lateral position of a vehicle $n$ in m \\
\hline
$v_n$ & The longitudinal velocity of a vehicle $n$ in m/s\\
\hline
$u_n$ & The lateral velocity of a vehicle $n$ in m/s\\
\hline
$a_n$ & The longitudinal acceleration of a vehicle $n$ in m/s$^2$\\
\hline
$A_\text{n}$ & The maximum longitudinal acceleration of a vehicle $n$ in m/s$^2$\\ 
\hline
$B_\text{n}$ & The maximum longitudinal deceleration of a vehicle $n$ in m/s$^2$\\ 
\hline
$e_n$ & The current lateral acceleration of a vehicle $n$ in m/s$^2$\\
\hline\hline
\end{tabular}
\caption{Descriptions and Notations}
\label{table:lc_4veh}
\end{table}
\end{center}

\section{Model} 
This section introduces the models we use for our prediction framework, which includes (1) a LSTM model to pre-train the longitudinal trajectories of neighboring vehicles, (2) a CF model to fit the acceleration of the initial 4 seconds of trajectories of the on-ramp vehicles and forecast their next 15 seconds of longitudinal trajectory, and (3) a RF classifier to classify LCs of on-ramp vehicles.

The reason why we do not use a LSTM model directly to forecast the central vehicles' trajectories is because, the insufficient number of on-ramp trajectories could make the result of the LSTM model lack accuracy. On the other hand, a CF model explains the traffic characteristics in a network of vehicles and can approximate their motions. Also, the errors of LSTM pre-training of neighbors' trajectories can sometimes be cancelled out by averaging the locations, velocities, and accelerations of the immediate leader and the nearest leader into those of an actual leader. Our model does not predict the lateral motion of the vehicles but rather uses a RF classifier to classify LCs as a binary process since the on-ramp generally has only one lane and the lateral motion during lane-keeping has relatively low contribution to the risk of collisions compared to the longitudinal motion. In fact, a number of studies have also approximated LCs as instantaneous processes \cite{gong2016optimal, 8855110}.

\subsection{\textbf{Framework}}

\begin{figure}
    \centering
    \includegraphics[width=0.8\linewidth]{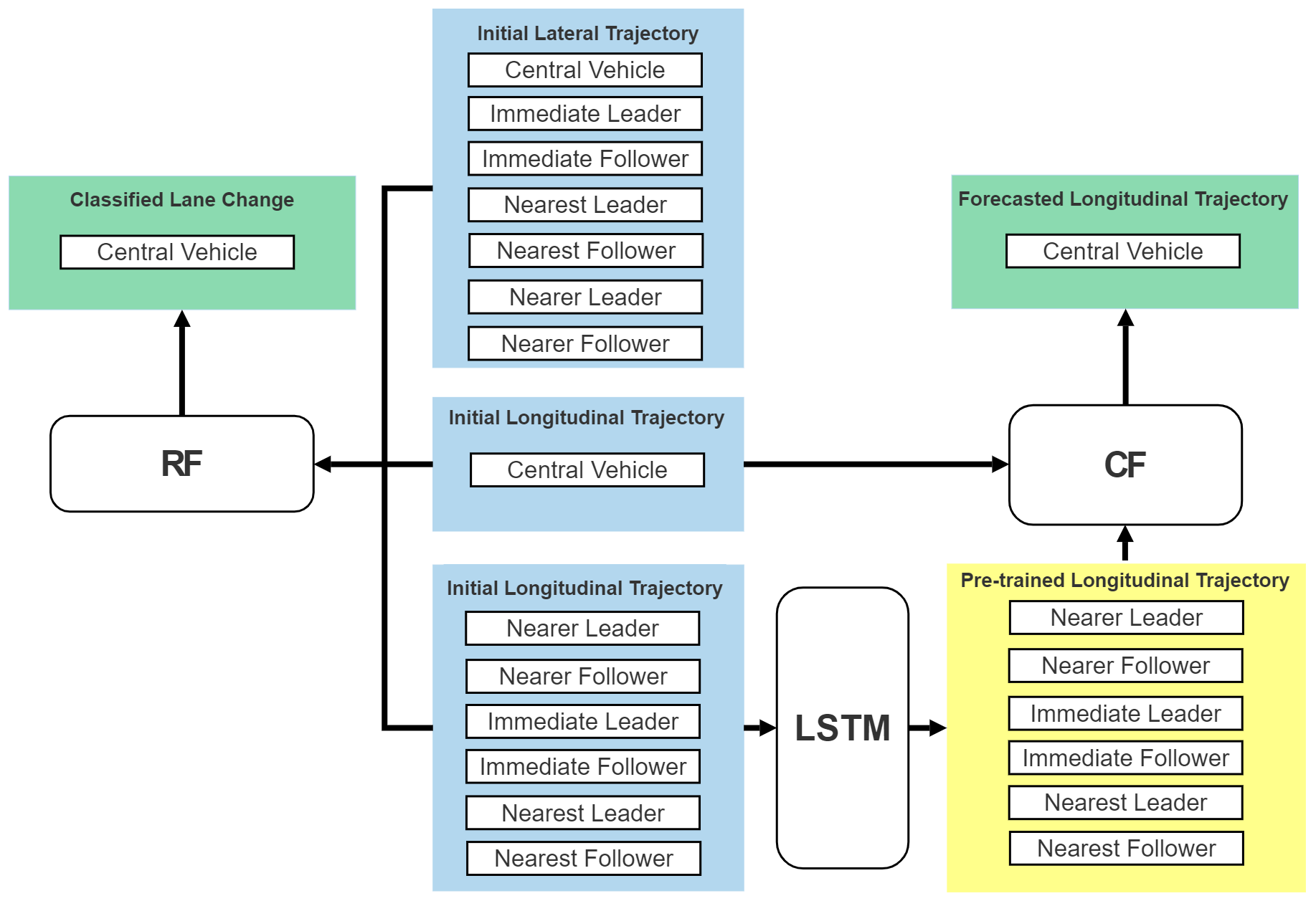}
    \caption{Prediction Framework}
    \label{frame}
\end{figure}

Fig.\ref{frame} shows our prediction framework. The given initial trajectories, the pre-trained trajectories are shown in blue and yellow respectively. A green background is used for classified LCs and forecasted trajectories.

We apply two LSTM models to pre-train the longitudinal locations of neighbors on the on-ramp and on the adjacent lane respectively. The reason we use two different models is because of the behavior differences of vehicles on the on-ramp and the adjacent lane. From the pre-trained trajectories of the neighbors, we calculate their velocities and accelerations.

Then, we fit a CF model based on the initial trajectory of the central vehicle and its actual leader. Given the pre-trained trajectories of the neighbors and the CF model, algorithm \ref{alg1} updates the longitudinal acceleration, velocity, and position for the next 15 seconds for the central vehicle. $\Delta t$ is the time step, $v_\text{max}$ is the maximum velocity, $F$ is a selected CF model, which outputs the acceleration value, and $I$ is the set of subscripts for neighbors in the adjacent lane. $x_i (0)$, $v_i (0)$, and $a_i (0)$ are the ending position, velocity, and acceleration of the central vehicle of the initial 4 seconds of the trajectory respectively.

At each time step, we extract the pre-trained trajectory of the 4 neighboring vehicles on the adjacent lane and the immediate leader from the 4th second of initial condition. Note that the nearest leader of the central vehicle can change in the 15 seconds of forecast. Therefore, we identify the new immediate leader and the nearest leader. If the pre-trained immediate leader's location is still less than the end position of the current lane, we calculate the average position, velocity, and acceleration of the pre-trained immediate leader and the pre-trained nearest leader, and set those as the information of the actual leader. Otherwise, the actual leader is set as the nearest leader.

Then, we calculate the acceleration of the central vehicle based on its last time step and the pre-trained actual leader. We also check if the value of the acceleration is acceptable. If the calculated value is higher than the maximum acceleration or lower than the maximum deceleration, we set it as the maximum acceleration or the maximum deceleration respectively. Then we update the velocity. Here, we make sure the velocity is between zero and the speed limit. By using a linear interpolation, we update the forecasted position of the central vehicle. This process continues until the 15th second of the forecast.

\begin{algorithm}
\caption{Central Vehicle Forecasted Position Update}\label{alg1}
\begin{algorithmic}[1]
\State $I:=\{l1, l2, f1, f2\}$
\State $t_\text{max}\leftarrow15$, $t\leftarrow 0$
\State $\hat{x}_i \leftarrow x_i (0)$, $\hat{v}_i \leftarrow v_i (0)$, $\hat{a}_i \leftarrow a_i (0)$
\While{$t<t_\text{max}$}
    \State Get $\tilde{x}_n $, $\tilde{v}_n $, $\tilde{a}_n $  $\forall  n \in I$
    \State Select $p \in I$ such that $ \tilde{x}_p-\hat{x}_i = \text{min} \{\tilde{x}_n-\hat{x}_i \mid \tilde{x}_n-\hat{x}_i>=0, n \in I\} $
    \If {$x_l \geq x_\text{end}$}
        \State $\tilde{x}_m=\tilde{x}_p$, $\tilde{v}_m=\tilde{v}_p$, $\tilde{a}_m=\tilde{a}_p$ 
    \Else
        \State $\tilde{x}_m=\frac{\tilde{x}_p+\tilde{x}_l}{2}$, $\tilde{v}_m=\frac{\tilde{v}_p+\tilde{v}_l}{2}$, $\tilde{a}_m=\frac{\tilde{a}_p+\tilde{a}_l}{2}$, 
    \EndIf
    \State $\hat{a}_i \leftarrow F(\hat{\Omega}_i, \tilde{\Omega}_m)$
    \If {$\hat{a}_{i}>A_{i}$}
        \State $\hat{a}_{i}\leftarrow A_{i}$
    \ElsIf {$\hat{a}_{i}<B_{i}$}
        \State $\hat{a}_{i}\leftarrow B_{i}$
    \EndIf 
    
    \State $\hat{v}_{i} \leftarrow \hat{v}_{i}+ \hat{a}_{i} \Delta t$
    
     \If {$\hat{v}_{i}>v_{\text{max}}$}
        \State $\hat{v}_{i}\leftarrow v_{\text{max}}$
    \ElsIf {$\hat{v}_{i}< 0$}
        \State $\hat{v}_{i}\leftarrow 0$
    \EndIf 
    
    \State $\hat{x}_{i} \leftarrow \hat{x}_{i} + \hat{v}_{i} \Delta t$, $t \leftarrow t+\Delta t$

\EndWhile

\end{algorithmic}
\end{algorithm}

The LC classification of the algorithm is a separate module from longitudinal forecast. Using the last time step of the initial trajectory, we predict whether the central vehicle is going to execute a LC in less then a certain number of seconds, or at an exact future time step, with a RF classifier.

\subsection{LSTM}
\begin{figure}
    \centering
    \includegraphics[width=0.8\linewidth]{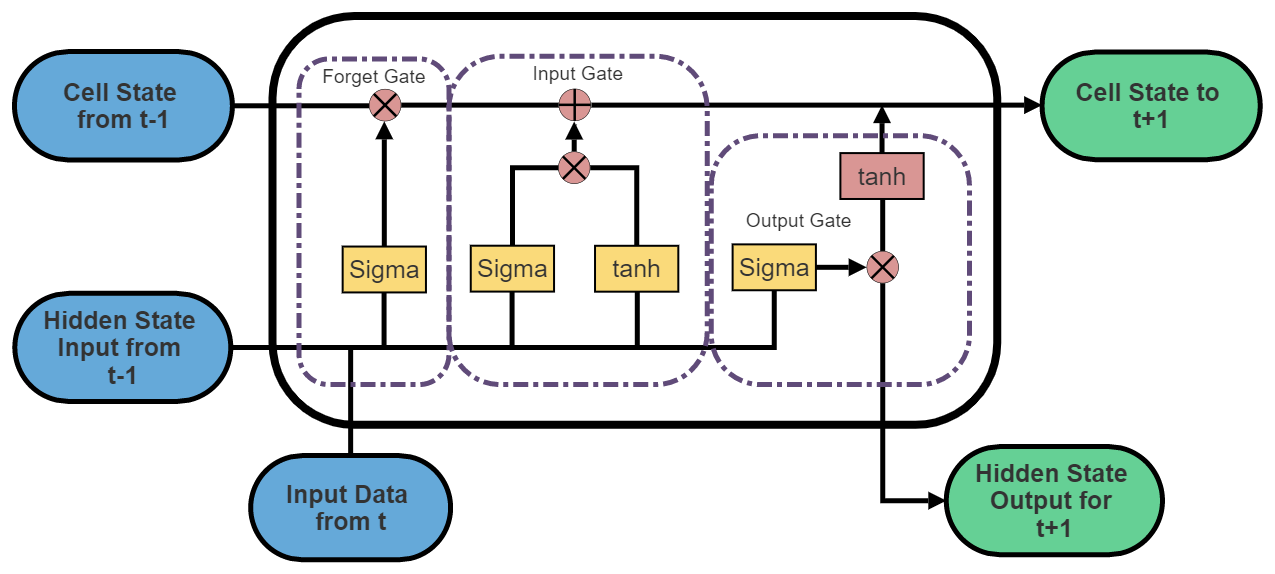}
    \caption{A LSTM Cell}
    \label{lstm_cell}
\end{figure}

LSTM is a kind of recurrent neural network that is robust in prediction of sequential data \cite{hochreiter1997long}. Fig.\ref{lstm_cell} shows a cell of LSTM network. Each cell takes the cell state and the hidden state from the last step $t-1$, and the current time step $t$ as the input. The inputs go through a forget gate which ignores irrelevant information, an input gate which acquires and stores new information, and an output gate which computes the output value for the next time step $t+1$ \cite{hochreiter1997long}. 

We use the LSTM network architecture recommended by Analytics Vidhya, which contains 2 encoding layers and 2 decoding layers with 100 neurons in each layer \cite{jagadeesh23_2020}. We choose Adam optimizer because it is an efficient algorithm for gradient descent for deep neural networks and is also easy to implement \cite{kingma2014adam}, and Huber loss function, because it is less susceptible to outliers than the mean-squared loss function \cite{huber1964robust}.

When training the LSTM for our prediction framework, we break the trajectories into sub-segments of 21 time steps, where each time step is 0.2 second. The first 20 time steps (4 seconds) are the input, and the last time step is the output.

When predicting the pre-trained trajectories of the neighbors, we use the first 4 seconds of real trajectories only. For example, when we predict the position of a neighbor at the 3rd second in the future, we use the last second of initial data and the first 3 seconds of the pre-trained data as the input sequence.

\subsection{CF Model}
We examine the application of 3 CF models: (1)Intelligent Driver Model (IDM), (2)Gipps' Model, and (3)Gazis-Herman-Rothery (GHR) Model, and compare their accuracy in forecasting the positions of the on-ramp vehicles.

\begin{enumerate}

\item{\textbf{IDM}}\\

IDM is a CF model that has relatively good performance in congested driving conditions\cite{treiber2000congested,chen2020modeling}, and is therefore expected to behave well for highway on-ramps. The desired inter-vehicle spacing is expressed as Eq.\ref{scur}.\\
\begin{equation}\label{scur}
\begin{split}
    {s}_{d}=s_0+h_d v_i +\frac{v_i (v_i - v_m)}{2\sqrt{A_i B_i}}
\end{split}
\end{equation}\\
where $s_d$ and $s_0$ are the desired and minimum inter-vehicle spacing for a central vehicle. To be specific, a vehicle can not move forward if its inter-vehicle spacing is below $s_0$. $h_d$ is the desired time headway. Under the IDM model, we express the acceleration of a vehicle as Eq.\ref{idm}.\\
\begin{equation}\label{idm}
\begin{split}
    a_i=A_i\left[1-{\left(\frac{v_i}{v_{d}}\right)}^{\delta}-{\left(\frac{{s}_{d}}{x_m-x_i}\right)}^{2}\right] 
\end{split}
\end{equation}\\
where $\delta$ is an exponent, and $v_d$ is the desired speed for a central vehicle.
In our prediction framework, the parameters to be determined for each vehicle are $s_0$, $h_d$, $A_i$, $B_i$, $v_d$, and $\delta$.\\

\item{\textbf{Gipps'}}\\
Gipps' model has shown characteristics of real traffic flow in simulation with synthetic data \cite{gipps1981behavioural}. It is modeled as Eq.\ref{gipps}.\\
\begin{equation} \label{gipps}
a_i=\alpha\frac{{(v_{m}-v_{i})}^{\beta}}{{(x_i-x_m)}^{\gamma}}
\end{equation}\\
where $\alpha$, $\beta$ and ${\gamma}$ are unit-less coefficients to be determined for each vehicle in our prediction framework. \\

\item{\textbf{GHR}}\\
According to GHR CF model, the acceleration of a central vehicle depends on the velocity of itself and the leader, and could be modeled by Eq.\ref{ghr_raw} \cite{brackstone1999car}. \\
\begin{equation}\label{ghr_raw}
a_i=\alpha v_i^{\beta} \frac{v_{m}-v_i}{{(x_{m}-x_i)}^{\gamma}}
\end{equation}\\
where $\alpha$, $\beta$ and ${\gamma}$ are unit-less coefficients to be determined for each vehicle in our prediction framework.\\

\end{enumerate}

To fit the CF models, we use the initial 4 seconds of longitudinal trajectories for both the central vehicle and its actual leader. We minimize the mean-squared error of the predicted acceleration for each initial trajectory.

\subsection{\textbf{RF Classifier}}

In our prediction framework, a RF classifier is applied to classify the LCs of central vehicles. A RF classifier performs training by bagging the samples into subsets and developing multiple decision trees \cite{ho1995random}. This method in general has a high classification accuracy and avoids the problem of overfitting \cite{ho1995random}. 

\begin{figure}
    \centering
    \includegraphics[width=0.55\linewidth]{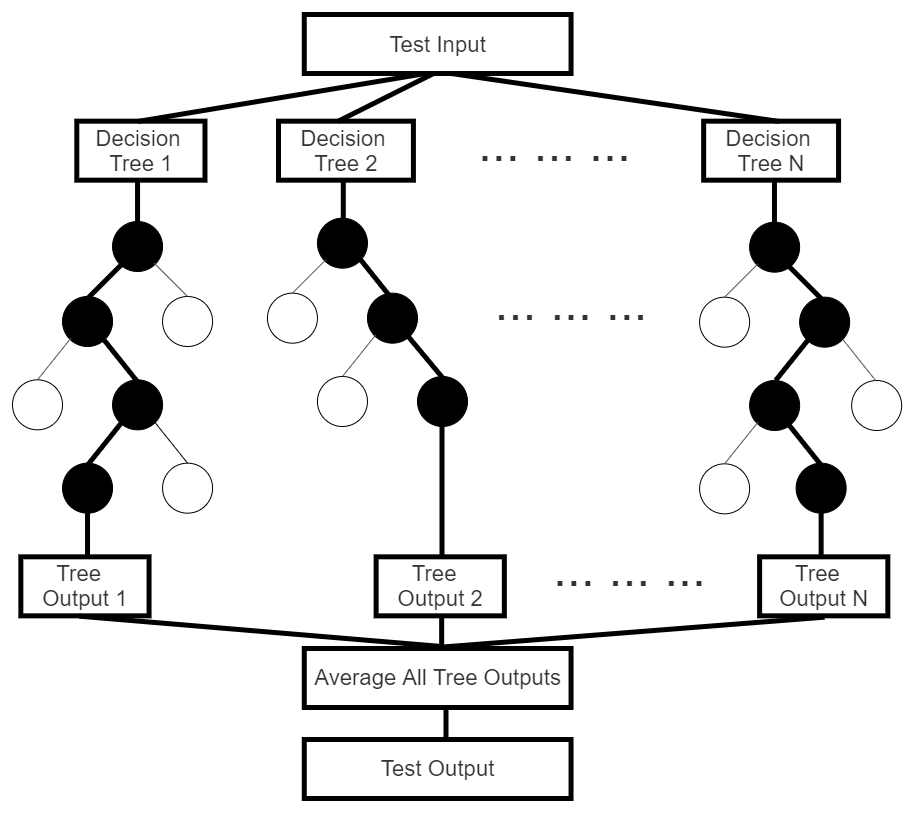}
    \caption{Random Forest Classifier Decision Making Process}
    \label{rf}
\end{figure}

Fig.\ref{rf} shows the decision making process of a RF classifier on a given test input. This classifier has $N$ decision trees, and the trees have different depth. Each decision tree yields a classification result for the input. Then, the results of all the decision trees are averaged to get the final output. 

We perform two types of LC classifications with RF classifiers:
\begin{enumerate}
    \item \textbf{Cumulative LC Classification} This type of LC prediction predicts whether a vehicle will execute a LC within the next certain number of seconds.
    \item \textbf{Exact LC Classification} This type of LC prediction predicts whether a vehicle will execute a LC at an exact time step.
\end{enumerate}

We develop a RF classifier for each classification. Since the time ranges from the 0th second to the 16th second, and there are two types of classification, there are 32 classifiers in total. The inputs of each classifier are listed by Eq.\ref{rf features}.\\

\begin{equation}\label{rf features}
\begin{aligned}
    R(t) := \underset{n \in \{l, l1, l2, f, f1, f2\} }{\cup} & \{|x_n (0)-x_i (0)|, y_n (0)-y_i (0), v_n (0), u_n (0), a_n (0), e_n (0)\}  \\ 
    \cup & \{{x}_i(0), {y}_i(0), {v}_i(0), {u}_i(0), {a}_i(0), {e}_i(0)\}
\end{aligned}
\end{equation}\\

where $R$ denotes the set of input features for the RF classifier. The features contain the velocity and acceleration vectors of a central vehicle and all of its neighbors. For example, $v_l (0)$ is the longitudinal velocity of the immediate leader. In addition, we include the longitudinal and lateral position of the central vehicles, and the longitudinal spacings and lateral distances between the neighbors and the central vehicle. In particular, $|x_n (0) - x_i (0)|$ and $y_n (0) - y_i (0)$ are the longitudinal spacing and lateral distance between one of its neighbors $\Omega_n$ and the central vehicle.

\section{Data Set}
\begin{figure}
    \centering
    \includegraphics[width=0.65\linewidth]{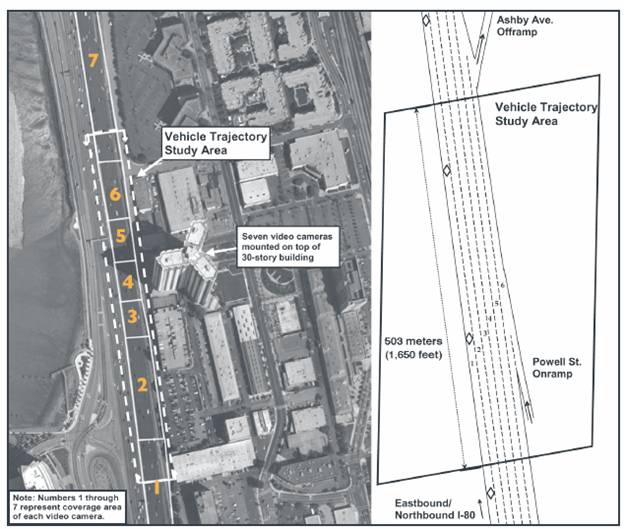}
    \caption{Geometry of I80 Freeway Data Collection Segment}
    \label{I80}
\end{figure}

In this study, we use the on-ramp vehicle trajectories from the NGSIM I80 data set \cite{alexiadis2006video}. Vehicle trajectories are collected from 7:50 am to 8:35 am on June 15, 2005, and are saved separately into three 15 min windows. In the rest of paper, we denote the three time windows 7:50am-8:05am, 8:05am-8:20am and 8:20-8:35 as time window 1, 2 and 3 respectively. Fig.\ref{I80} shows the geometry of the segment of freeway. The total length of the segment of freeway is 503m, and the on-ramp is attached to the basic segment of freeway from 170m to 230m, where on-ramp vehicles execute MLCs. Each vehicle has a unique ID, and the data set provides longitudinal and lateral position, speed, acceleration, as well as the lane ID every 0.1 second. However, there are some skipped time steps. To address this issue, we use a linear interpolation to fill in the missing positions. In addition, the NGSIM data set has huge errors in its velocities and accelerations \cite{coifman2017critical, thiemann2008estimating, altche2017lstm}. Therefore, we apply a Savitzky-Golay filter to smoothen the longitudinal and lateral position, and then recalculate the velocity and acceleration vectors by using a second-order interpolation\cite{jiang2019trajectory}. 

We extract the trajectories of vehicles every 0.2 second. In all, there are 713 on-ramp vehicles in I80 data set, and 385 of then have a trajectory of over 19 seconds. Among the 385 vehicles, 117 of them are from time window 1, 129 from time window 2, and 139 from time window 3. For each of those vehicles, we extract their trajectories along with the information of the 6 neighbors.

In situations where some neighbors are missing, we place a virtual vehicle at a specific longitudinal position at the center of its lane. If the vehicle missing is an immediate leader, we put it at the end of the on-ramp. We put the virtual vehicle at 500m abd and -500m for all other missing leaders and followers respectively. We also set their velocity and acceleration vectors to zero for convenience. In addition, we also extract the longitudinal trajectories of vehicles on the adjacent lane in the ramp influence area for the pre-training. In all, there are 1529 such trajectories on the adjacent lane for the three time windows.

We use time window 1 and 3 for training, and time window 2 for testing. The training set is used to perform training for LSTM models and RF classifiers. For LSTM pre-training, we develop separate models for vehicles on the on-ramp and in the adjacent lane. For the training RF classifiers, we construct 16 training sets each for cumulative and exact LC classifications. 
For the cumulative LC classification, the positive samples are collected at a time when the vehicle will execute a LC in less than $t$ seconds, and the negative samples more than $t$ seconds. For the exact LC classification, the positive samples are collected when the vehicle will execute a LC in exactly $t$ seconds, and the negative samples more than $t$ seconds. We make both types of training sets for all $t$'s between 0 to 15. To make sure the samples are even, we collect one positive sample and one negative sample for each vehicle when making each training set.

For the testing set, we randomly extract two sub-trajectories of 19 seconds together with their 6 neighbors for each central vehicle. The first 4 seconds of each sub-trajectory are the test input, and the rest are the output. The initial trajectories are also used to fit with a CF model for the accelerations. Then we forecast the central vehicles' next 15 seconds. To test the LC classifications, we partition the testing set in a similar way as the training set.

\section{Results} 
This section discusses the results of the LSTM pre-training, the CF parameters, the longitudinal trajectory forecast, and the LC classification.

\subsection {LSTM Pre-training}
\begin{figure}
     \centering
     \begin{subfigure}[h]{0.49\textwidth}
         \centering
         \includegraphics[width=1\textwidth]{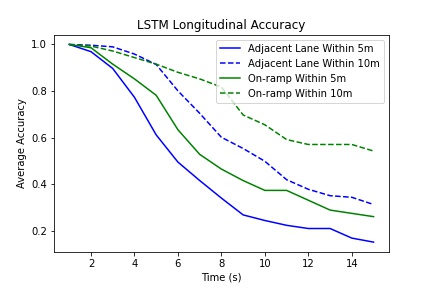}
         \caption{Accuracy Rate of Pre-training}
         \label{pt_p}
     \end{subfigure}
     \begin{subfigure}[h]{0.49\textwidth}
         \centering
         \includegraphics[width=1\textwidth]{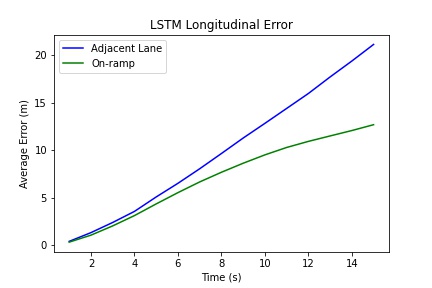}
         \caption{Average Error of Pre-training}
         \label{pt_err}
     \end{subfigure}
     \caption{LSTM Pre-training Result}
     \label{pt}
\end{figure}

We first examine the pre-training results of LSTM. In particular, we present the performance of the LSTM models on predicting the positions of neighbors of randomly extracted central vehicle trajectories in the testing set. Fig.\ref{pt} shows the results of the LSTM models on predicting the longitudinal positions of vehicles. The blue and green lines indicate pre-training on the adjacent lane and the on-ramp respectively. Fig.\ref{pt_p} shows the accuracy of the models. The dashed lines and the solid lines indicate the probabilities where the predicted positions lie within 10m and 5m of the actual longitudinal positions for every second of prediction respectively. Fig.\ref{pt_err} shows the average distance deviations from the actual longitudinal positions for both lanes. 

For both of the lanes, the prediction accuracy of the LSTM models decreases as the time increases. The prediction of neighbors on the on-ramp has a slightly better result. However, neither of the LSTM models provide a promising accuracy. However, we will show how the implementation of a CF model can mitigate the error of the longitudinal trajectory forecast of central vehicles compared with using LSTM to pre-train their neighbors.

\subsection{CF Model Fitting}

We use the Scipy package to fit the acceleration of the initial 4 seconds of the trajectory samples in time window 2, and minimize their mean-square errors \cite{virtanen2020scipy}. We also restrict each fitting parameter to a specific range. Table \ref{table:idm}, \ref{table:gipps} and \ref{table:ghr} show the fitting results under IDM, Gipps', and GHR model respectively. In each table, we include the range of fitting and the distribution of each parameter, as well as the distribution of the mean-squared errors of the fitted accelerations for all the initial trajectories.

\begin{center}
\begin{table}[]
\centering
\begin{tabular}{||p{2.5cm} | p{3cm} || p{1.5cm} | p{1.5cm} |p{4cm} ||}
\hline\hline
 \textbf{Parameter} &  \textbf{Range of Fitting} &\textbf{Mean} & \textbf{Median} & \textbf{Standard Deviation}  \\
\hline\hline
 $s_0$ (m) & [5, 30] & 10.68 & 6.92 & 8.47\\
\hline
 $h_d$ (s) & [0.5, 6] & 1.79 & 0.91 & 1.88\\
\hline
 $A_i$ (m/s$^{2}$)& [0.5, 5] & 3.34  & 3.75 & 1.64\\
\hline
 $B_i$ (m/s$^{2}$)& [0.5, 5]&2.63 & 1.63 & 2.10\\
\hline
 $v_d$ (m/s$^{2}$) & [5, 35] &18.19 & 7.62 & 13.89\\
\hline
 $\delta$ & [0, 10] & 5.19 & 4.48 & 4.04\\
\hline
Mean-Squared Error (m$^2$/s$^{4}$) & N/A & 0.08 & 0.02 & 0.26\\
\hline\hline
\end{tabular}
\caption{IDM Model Parameters Distribution}
\label{table:idm}
\end{table}
\end{center}
\begin{center}
\begin{table}[]
\centering
\begin{tabular}{||p{2.5cm} | p{3cm} || p{1.5cm} | p{1.5cm} |p{4cm} ||}
\hline\hline
 \textbf{Parameter} &  \textbf{Range of Fitting} &\textbf{Mean} & \textbf{Median} & \textbf{Standard Deviation}\\
\hline\hline
 $\alpha$ & [-10, 10] & 2.26 & 0.00 & 5.62 \\
\hline
 $\beta$ & [-5, 5] &0.76 & 0.70  & 2.42  \\
\hline
 $\gamma$ & [-5, 5] & -1.42 & -0.95 & 3.09 \\
\hline
Mean-Squared Error (m$^2$/s$^{4}$) & N/A& 0.18  & 0.06 & 0.29 \\
\hline\hline
\end{tabular}
\caption{Gipps' Model Parameters Distribution}
\label{table:gipps}
\end{table}
\end{center}
\begin{center}
\begin{table}[]
\centering
\begin{tabular}{||p{2.5cm} | p{3cm} || p{1.5cm} | p{1.5cm} |p{4cm} ||}
\hline\hline
 \textbf{Parameter} &  \textbf{Range of Fitting} &\textbf{Mean} & \textbf{Median} & \textbf{Standard Deviation} \\
\hline\hline
 $\alpha$ & [-10, 10] & 2.84  & 2.14 & 4.77 \\
\hline
 $\beta$ & [-5, 5] & 0.06 & -0.16  & 3.89 \\
\hline
 $\gamma$ & [-5, 5] &  0.96 & 0.9 & 1.62\\
\hline
Mean-Squared Error (m$^2$/s$^{4}$) & N/A & 0.12 & 0.06 & 0.22 \\
\hline\hline
\end{tabular}
\caption{GHR Model Parameters Distribution}
\label{table:ghr}
\end{table}
\end{center}
Fitting results show that all three models yield small average mean-squared error on the initial trajectories. However, some fitted parameters have high standard deviations. One reason is that the CF models may not explain all initial conditions of vehicles, which makes some fitted parameters to take the boundary values. This is also explained by a high standard deviation of the mean-square error, which means a skewed distribution caused by some poor fittings.

However, comparing the mean of our fitted parameters with the existing studies, we find similar values\cite{chen2020modeling, brackstone1999car, ciuffo2012thirty}.

\subsection {Longitudinal Forecast}
\begin{figure}
     \centering
     \begin{subfigure}[h]{0.49\textwidth}
         \centering
         \includegraphics[width=1\textwidth]{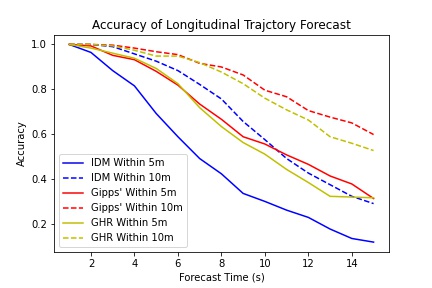}
         \caption{Accuracy Rate of Forecast}
         \label{fr_p}
     \end{subfigure}
     \begin{subfigure}[h]{0.49\textwidth}
         \centering
         \includegraphics[width=1\textwidth]{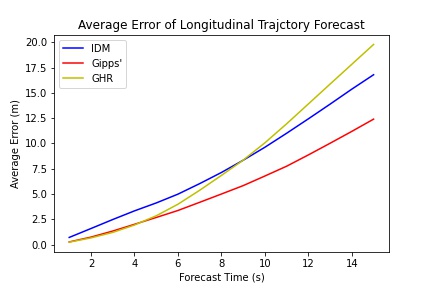}
         \caption{Average Error of Forecast}
         \label{fr_err}
     \end{subfigure}
     \caption{Longitudinal Forecast Result of Central Vehicles}
     \label{fr}
\end{figure}

Fig.\ref{fr} shows the results of the forecast of longitudinal positions of central vehicles with the three kinds of CF models on the testing data set. In Fig.\ref{fr_p}, The solid and dashed lines indicate the probabilities where the forecasted longitudinal position is less than 5m and 10m away from the real position respectively. Fig.\ref{fr_err} shows the average error of the forecast.

In terms of the accuracy rate, the Gipps' model and GHR model yields similar results, and much better than that of IDM model. Fig.\ref{fr_p} indicates that at times before the 8th second and the 5th second, the forecasted longitudinal positions lies within 10m and 5m from the real positions with probabilities higher than 90\% respectively, for both Gipps' model and GHR model. However, the IDM model yields significantly lower accuracy. This could be because the IDM model has more parameters, and over-fitting may more likely exist with only the 20 initial data points. Comparing the average error of forecast in Fig.\ref{fr_err}, we can see that GHR has yielded a highest average error for times later than the 9th second. This could be explained by some extremely inaccurate fittings.

\begin{figure}
     \centering
     \begin{subfigure}[h]{0.49\textwidth}
         \centering
         \includegraphics[width=1\textwidth]{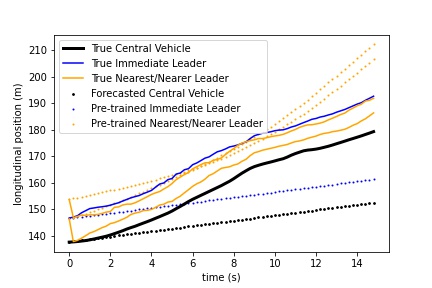}
         \caption{}
         \label{fw}
     \end{subfigure}
     \begin{subfigure}[h]{0.49\textwidth}
         \centering
         \includegraphics[width=1\textwidth]{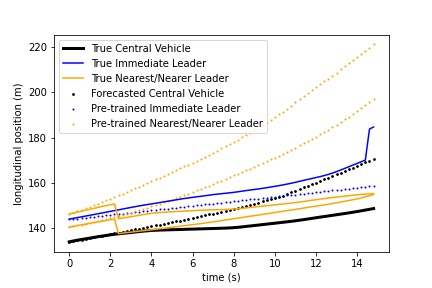}
         \caption{}
         \label{bw}
     \end{subfigure}
     \caption{Traffic Shock Waves}
     \label{sw}
\end{figure}

Then, we analyse circumstances where the longitudinal trajectory forecast is inaccurate. One key deficiency of our model is when a traffic shock wave exists in the near future. For example, Fig.\ref{fw} and Fig .\ref{bw} show the two situations where shock waves start sometime during the forecast. A jump or discontinuity in the curves represent a change in neighbor. In Fig.\ref{fw}, the central vehicle has experienced a forward-propagating wave at the 2nd second, and then a backward-propagating wave at around the 9th second. In Fig.\ref{bw}, the central vehicle is travelling in a traffic jam, where it nearly stops between the 3rd second and the 7th second. However, the prediction model fails to anticipate these situations. The external conditions such as a traffic break down or an end of the traffic jam could possibly be detected by observing the vehicles further downstream, instead of the 6 neighboring vehicles.

\begin{figure}
     \centering
     \begin{subfigure}[h]{0.49\textwidth}
         \centering
         \includegraphics[width=1\textwidth]{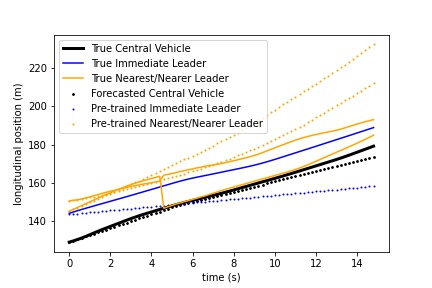}
         \caption{}
         \label{gd1}
     \end{subfigure}
     \begin{subfigure}[h]{0.49\textwidth}
         \centering
         \includegraphics[width=1\textwidth]{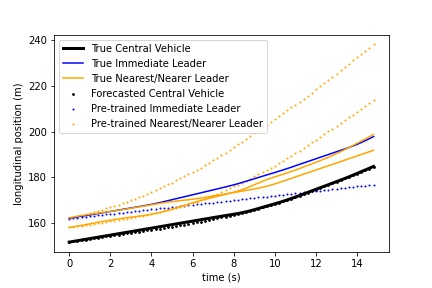}
         \caption{}
         \label{gd2}
     \end{subfigure}
     \caption{Accurate Forecast Result with Inaccurate Pre-training}
     \label{good_result}
\end{figure}

However, the result shows that our model has an advantage in forecasting the longitudinal positions of the central vehicles even in situations where the LSTM model has relatively inaccurate pre-training result for the neighbors. Fig.\ref{good_result} shows examples where the forecasted trajectory of the central vehicle is accurate throughout the 15 seconds of forecast while the LSTM models does poor prediction for the position of neighbors. In situations shown in Fig.\ref{gd1} and Fig.\ref{gd2}, the LSTM underestimates the speed of the neighbors on the on-ramp, and overestimates the speed of neighbors on the adjacent lane. However, by taking the average position of the nearest leader and the immediate leader, we still yield a valid estimation of the actual leader's position, and thereby producing an accurate forecasts for the central vehicle. 

Comparing the result of the trajectory forecast in Fig.\ref{fr} with the LSTM pre-training on the neighbors on the on-ramp in Fig.\ref{pt}, we find that the accuracy has increased significantly for all time steps with the application of GHR model or Gipps' model.

\subsection {LC Classification}

\begin{figure}
     \centering
     \begin{subfigure}[h]{0.49\textwidth}
         \centering
         \includegraphics[width=1\textwidth]{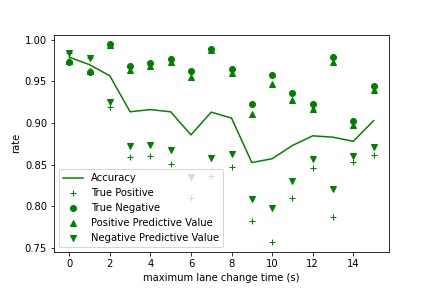}
         \caption{Cumulative LC Prediction}
         \label{cum_lc}
     \end{subfigure}
     \begin{subfigure}[h]{0.49\textwidth}
         \centering
         \includegraphics[width=1\textwidth]{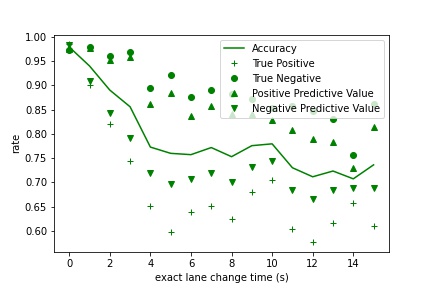}
         \caption{Exact LC Prediction}
         \label{spc_lc}
     \end{subfigure}
     \caption{RF Classifier LC Prediction Result}
     \label{lc_pred}
\end{figure}

Fig.\ref{lc_pred} show the metrics of the result of the both cumulative and exact LC classification with RF classifiers on the testing data set.

\begin{figure}
     \centering
     \begin{subfigure}[b]{0.49\textwidth}
         \centering
         \includegraphics[width=1\textwidth]{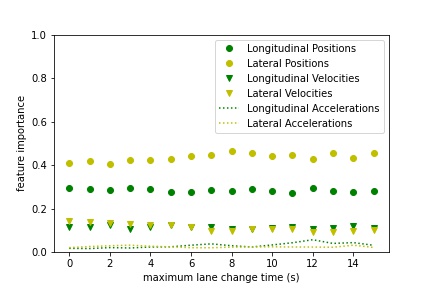}
         \caption{Cumulative LC Prediction: Feature Importance by Spacial Features   }
         \label{cum_pos}
     \end{subfigure}
     \begin{subfigure}[b]{0.49\textwidth}
         \centering
         \includegraphics[width=1\textwidth]{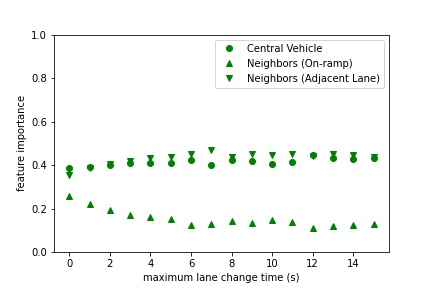}
         \caption{Cumulative LC Prediction: Feature Importance by Vehicles}
         \label{cum_veh}
     \end{subfigure}
     \centering
     \begin{subfigure}[b]{0.49\textwidth}
         \centering
         \includegraphics[width=1\textwidth]{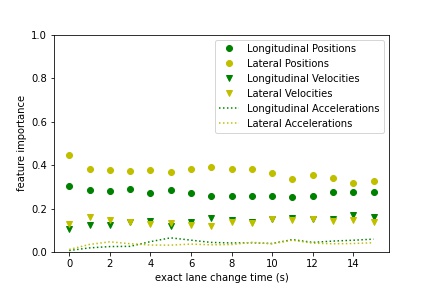}
         \caption{Exact LC Prediction: Feature Importance by Spacial Features}
         \label{spc_pos}
     \end{subfigure}
     \begin{subfigure}[b]{0.49\textwidth}
         \centering
         \includegraphics[width=1\textwidth]{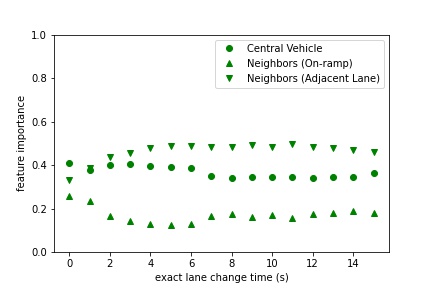}
         \caption{Exact LC Prediction: Feature Importance by Vehicles}
         \label{spc_veh}
     \end{subfigure}
     \caption{RF Classifier Feature Importance}
     \label{fi}
\end{figure}

According to Fig.\ref{cum_lc}, LC prediction is relatively accurate for cumulative LC prediction, and is in general over 90\% for most of the time steps. There is also a trend that the accuracy decreases with time first and then increases. The true negative rate and the positive predictive rate are higher than the total accuracy and is on average higher than 95\%. This means that it is less likely for the classifier to miss-classify a lane-keeping sample into a LC sample than vise versa. According to Fig.\ref{spc_lc}, exact LC prediction does not have very promising result as its accuracy falls below 85\% from the 4th second to the 15th second. This is also not surprising because the negative training samples are randomly collected. For example, when training the LC at the 14th second, the samples that execute a LC at the 15th second is also labeled negative, and this small difference of time to execute a LC in the future may not be detected by the current information.

Fig.\ref{fi} shows the feature importance of the two types of LC classification. Fig.\ref{cum_pos} and Fig.\ref{spc_pos} show the feature importance by dimensional features, including positions, velocities, and accelerations of all the vehicles. For both types of classifications, the lateral positions of the central vehicles and the neighbors are the most important features. In general, positions are more important than velocities, and velocities are more important than acceleration. Fig.\ref{cum_veh} and Fig.\ref{spc_veh} show the feature importance of features from the central vehicle, neighbors on the on-ramp and in the adjacent lane. According to the RF classifiers, the neighbors on the adjacent lane influence the decisions to execute LCs the most, followed by the central vehicle and the neighbors on the on-ramp. 

\section{Conclusions and Discussions}
This paper aims to forecast the longitudinal trajectories and LCs of vehicles on the on-ramp. We propose a prediction framework that contains a LSTM model to pre-train the positions of the neighbors, a CF model to forecast the longitudinal trajectories, and a RF classifier to classify LCs of the central vehicles. 

Our results indicate that the application of a Gipps' or GHR CF model has yielded high accuracy for first 5 seconds of longitudinal trajectory forecast, and also significantly mitigates the error when the LSTM pre-training is relatively inaccurate. The CF models also produce reasonable parameters. In addition, We find appealing results for cumulative LC predictions. However, Our model fails in situations where a traffic wave exists during the forecast time horizon, and is not capable of providing the precised time of a LC that is more than 3 seconds in the future. 

A key difficulty for our research is the lack of data for on-ramps. This is a bottleneck for improving prediction accuracy and generalizability to on-ramp locations other than I80. In the future, we will collect more data near highway on-ramps, and use a more accurate machine learning model to pre-train the trajectories to further improve the accuracy.

\section{Contribution}
The authors confirm contribution to the paper as follows: study conception and
design: Nachuan Li, Riley Fischer, Wissam Kontar; data collection: Nachuan Li; analysis and
interpretation of results: Nachuan Li, Riley Fischer, Wissam Kontar; draft manuscript
preparation: Nachuan Li, Riley Fischer, Wissam Kontar, Soyoung Ahn. All authors have reviewed the results and approved
this version of the manuscript. 

\section{Acknowledgement}
The authors sincerely thank Yongle Yuan, Chaolun Xu, Austen Z. Fan, and Ron Yang for their kind support and helpful discussions.

\newpage

\bibliographystyle{trb}
\bibliography{trb_template}
\end{document}